\documentclass[conference,compsoc]{IEEEtran}
\ifCLASSOPTIONcompsoc
  \usepackage[nocompress]{cite}
\else
  \usepackage{cite}
\fi
\ifCLASSINFOpdf
\else
\fi
\usepackage{amssymb}
\usepackage{amsmath}
\usepackage{mathabx}
\usepackage{stmaryrd}
\usepackage{graphicx}
\usepackage{booktabs}

\DeclareMathAlphabet\mathcalbf{OMS}{cmsy}{b}{n}
\DeclareMathOperator{\vectorize}{\textit{vec}}
\DeclareMathOperator{\matricize}{\textit{mat}}
\DeclareMathOperator{\degreemat}{\textit{degree}}

\newtheorem{example}{Example}
\newtheorem{remark}{Remark}

\begin{document}

%
\title{Graph Tensor Networks: An Intuitive Framework for Designing \\ Large-Scale Neural Learning Systems on Multiple Domains}


\author{

    \IEEEauthorblockN{Yao Lei Xu, Kriton Konstantinidis, Danilo P. Mandic}
    \IEEEauthorblockA{Department of Electrical and Electronic Engineering \\ 
    Imperial College London \\
    London SW7 2AZ, UK \\
    Email: \{yao.xu15, k.konstantinidis19, d.mandic\}@imperial.ac.uk}


    
    
}


%


\maketitle

\begin{abstract}
Despite the omnipresence of tensors and tensor operations in modern deep learning, the use of tensor mathematics to formally design and describe neural networks is still under-explored within the deep learning community. To this end, we introduce the Graph Tensor Network (GTN) framework, an intuitive yet rigorous graphical framework for systematically designing and implementing large-scale neural learning systems on both regular and irregular domains. The proposed framework is shown to be general enough to include many popular architectures as special cases, and flexible enough to handle data on any and many data domains. The power and flexibility of the proposed framework is demonstrated through real-data experiments, resulting in improved performance at a drastically lower complexity costs, by virtue of tensor algebra.
\end{abstract}


%
\IEEEpeerreviewmaketitle

\section{Introduction}


Modern neural networks are effectively comprised by a cascade of tensor operations that are interleaved with non-linear activation functions. However, despite the omnipresence of tensors in deep learning research and popular software libraries, the use of tensor mathematics to formally describe and design neural learning systems is not yet common practice within the research community. 

Tensors are a multi-dimensional generalization of vectors (order-1 tensors) and matrices (order-2 tensors). Tensor operations exploit the multi-dimensional structure inherent to Big Data applications to efficiently operate on large-dimensional data at low complexity, while preserving their inherent structure and interpretability \cite{TDforSP}. In this way, tensors help bypass the bottlenecks imposed by the Curse of Dimensionality, reducing the associated computational costs from an exponential one to a linear one in the data dimensions \cite{cichocki2016tensor, Mandic2017}. Despite the obvious advantages, the multi-linear algebra, a branch of mathematics dealing with tensors, is not as intuitive as the classical linear algebra. In addition, tensor expressions are not as easily manipulated, and the notation can be overwhelming due to the necessity of indexing operations across multiple dimensions. As a remedy to these conceptual obstacles, graphical approaches known as Tensor Networks (TN) have been developed in the quantum physics community, with the aim to illustrate and implement complex tensor operations as mathematical graphs, in an insightful and mathematically rigorous manner \cite{orus2019tensor, pan2020contracting, kisil2022accelerating}. The design of neural networks would therefore greatly benefit from the TN framework, as it would allow for the visualization and manipulation of large-scale, multi-dimensional operations through simple graphical illustrations. 

To this end, we introduce a TN based framework for describing neural networks through tensor mathematics. This allows for an intuitive and rigorous way of designing neural architectures for large- and multi-dimensional data across any and many data domains. The proposed framework is shown to be general enough to include many modern neural network architectures as special cases. Furthermore, it opens up avenues for the inclusion of sophisticated tensor techniques developed in quantum many-body physics in order to design new classes of neural networks, which are not possible to arrive at using their current design.

The rest of the paper is organised as follows. Related work is first discussed in Section \ref{sec:related_work}. The mathematical preliminaries necessary to follow this work are presented in Section \ref{sec:prelim}. The proposed Graph Tensor Network (GTN) framework is introduced in Section \ref{sec:gtn}. Several classical neural network architectures, such as Dense, Convolutional, Graph, Attention, and Recurrent Neural Networks, are shown to be special cases of the proposed framework, as elaborated in Section \ref{sec:classical_nnas}. Section \ref{sec:beyond} discusses the application of advanced tensor techniques for improving classical neural networks. The power and flexibility of the proposed framework is then demonstrated through real-data experiments in Section \ref{sec:experiments}. Finally, Section \ref{sec:conclusion} concludes the approach and summarizes the findings. 

\section{Related Work} \label{sec:related_work}

The joint consideration of graphs, tensors, and neural networks offers enormous potential and was first discussed in \cite{xu2020recurrent, xu2020multigraph}. This was achieved by leveraging on the concept of graph filters and tensor networks to develop low-complexity neural networks for time-domain modelling. Despite success, this initial work was highly specialized, as the motivation arose from the application domain rather than aiming to solve a general learning paradigm. The present work instead builds on this previous work by extending the considered framework to handle tensor-variate data on any and many domains, which is general enough to include numerous classical neural network architectures as special cases. The use of Tensor Decomposition (TD) techniques to reduce the complexity of neural networks was also explored for many architectures, such as Dense Neural Networks \cite{ttnn, calvi2019compression}, Convolutional Neural Networks \cite{sun2020deep}, Recurrent Neural Networks \cite{ttrnn, xu2021ttrnn}, and the Attention Mechanism \cite{xu2022attention}. However, the existing work was primarily concerned with the compression of neural networks, while this paper introduces a general framework for describing neural networks under one unified umbrella of tensor networks. Authors in \cite{cohen2016expressive} were the first to analyze the expressive power of convolutional neural networks through tensor networks. However, instead of considering one particular architecture, the framework proposed here applies more generally across many existing architectures. The idea of multi-graph modelling via neural networks was also discussed in \cite{monti2017geometric, geng2019spatiotemporal}, however, it did not leverage on the graphical method of tensor networks.

\section{Preliminaries} \label{sec:prelim}

This section introduces the mathematical preliminaries necessary to follow this work. We refer the readers to \cite{cichocki2016tensor, Mandic2017} for an in-depth treatment of tensor algebra, and \cite{stankovic2019graph, stankovic2019graphII, stankovic2020graphIII} for data analytics on graphs. 

\subsection{Basic Tensor Algebra}

\subsubsection{Definitions} 

An order-$N$ tensor (denoted by a bold calligraphic letter), $\mathcalbf{A} \in \mathbb{R}^{I_1 \times \cdots \times I_N}$, is a multi-dimensional array with $N$ modes (dimensions), where the $n$-th mode is of size $I_n$, $n=1, \ldots, N$. A matrix (denoted by a bold uppercase letter), $\mathbf{A} \in \mathbb{R}^{I_1 \times I_2}$, is an order-$2$ tensor. A vector (denoted by a bold lowercase letter), $\mathbf{a} \in \mathbb{R}^{I_1}$, is an order-$1$ tensor. A scalar (denoted by a lowercase letter), ${a} \in \mathbb{R}$, is an order-$0$ tensor. The $(i_1, \ldots, i_N)$-th entry of an order-$N$ tensor is denoted by $a_{i_1, \cdots, i_N} \in \mathbb{R}$, where $i_n = 1, \ldots, I_n$ for $n=1,\ldots,N$.

\subsubsection{Vectorization}

\begin{figure}[h]
    \centering
    \includegraphics[width=0.8\linewidth]{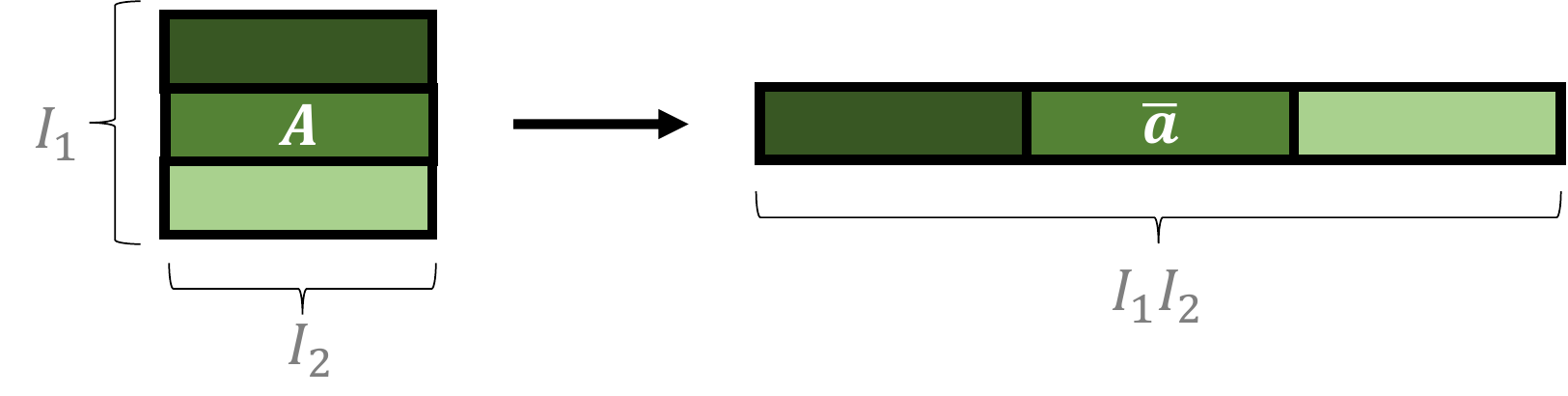}
    \caption{Vectorization of an order-$2$ tensor.}
    \label{fig:vectorization}
\end{figure}

An order-$N$ tensor, $\mathcalbf{A} \in \mathbb{R}^{I_1 \times \cdots \times I_N}$, can be vectorized to generate a long vector, $\vectorize \left( \mathcalbf{A} \right) = \mathbf{\bar{a}} \in \mathbb{R}^{I_1 \cdots I_N}$. The inverse operation is referred to as the vector tensorization. Figure \ref{fig:vectorization} illustrates the vectorization operation for an order-$2$ tensor.

\subsubsection{Matricization}

\begin{figure}[h]
    \centering
    \includegraphics[width=0.8\linewidth]{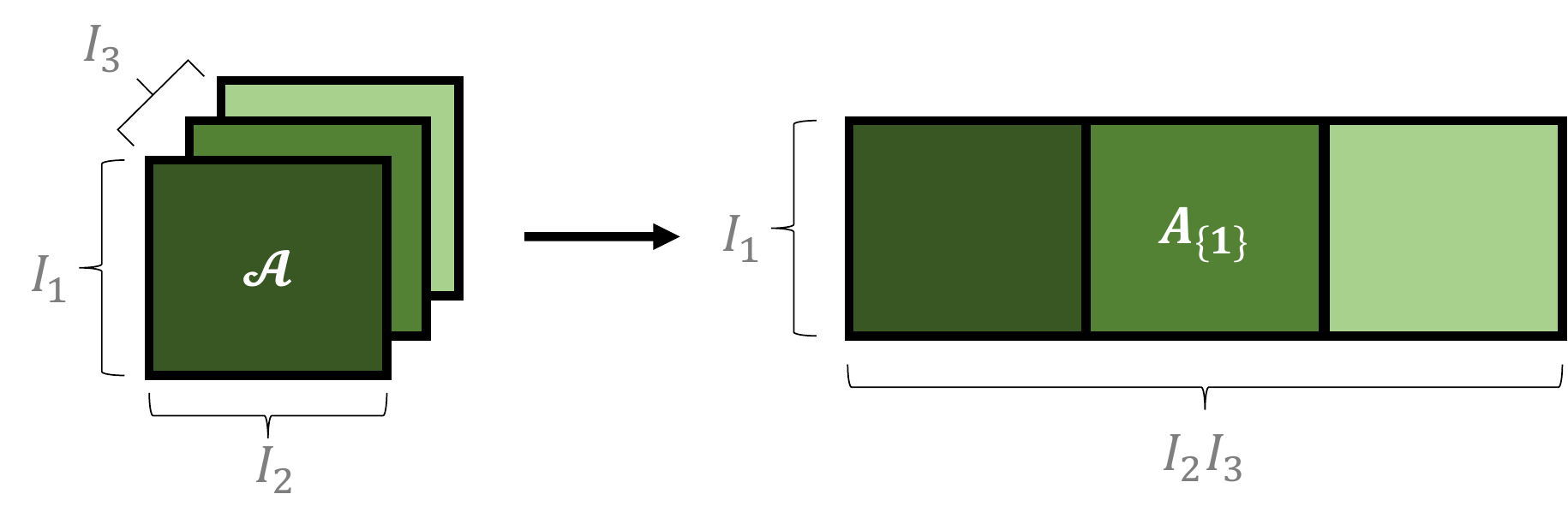}
    \caption{Matricization of an order-$3$ tensor with respect to mode-$1$.}
    \label{fig:matricization}
\end{figure}

An order-$N$ tensor, $\mathcalbf{A} \in \mathbb{R}^{I_1 \times \cdots \times I_N}$, can be matricized along its $n$-th mode to generate the matrix, $\matricize \left( \mathcalbf{A}, n \right) = \mathbf{A}_{\{n\}} \in \mathbb{R}^{I_n \times (I_1 \cdots I_{n-1} I_{n+1} \cdots I_N)}$, where the subscript, $\{n\}$, denotes the mode of matricization. The inverse operation is referred to as the matrix tensorization. Figure \ref{fig:matricization} illustrates the matricization operation for an order-$3$ tensor.

\subsubsection{Kronecker Product}

A (left) Kronecker product of the form, $\textbf{C} = \textbf{A} \otimes \textbf{B}$, between two matrices, $\textbf{A} \in \mathbb{R}^{I_1 \times I_2}$ and $\textbf{B} \in \mathbb{R}^{J_1 \times J_2}$, yields a block matrix, $\textbf{C} \in \mathbb{R}^{I_1 J_1 \times I_2 J_2}$, as
\begin{equation}
    \textbf{C} = 
    \begin{bmatrix}
    a_{1, 1} \textbf{B} & \cdots & a_{1, I_2} \textbf{B} \\
    \vdots & \ddots & \vdots \\
    a_{I_1, 1} \textbf{B} & \cdots & a_{I_1, I_2} \textbf{B} \\
    \end{bmatrix}
\end{equation}



\subsubsection{Tensor Contraction} 

An $(n, m)$-tensor contraction of the form, $\mathcalbf{C} = \mathcalbf{A} \times_n^m \mathcalbf{B}$, between an order-$N$ tensor, $\mathcalbf{A} \in \mathbb{R}^{I_1 \times \dots \times I_N}$, and an order-$M$ tensor, $\mathcalbf{B} \in \mathbb{R}^{J_1 \times \dots \times J_M}$, over the $n$-th mode of $\mathcalbf{A}$ and the $m$-th mode of $\mathcalbf{B}$, where $I_n=J_m$, results in an order-$(N+M-2)$ tensor, $\mathcalbf{C} \in \mathbb{R}^{I_1 \times \dots \times I_{n-1} \times I_{n+1} \times \cdots J_1 \times J_{m-1} \times J_{m+1} \cdots J_M}$, with the entries $c_{i_1,\dots,i_{n-1},i_{n+1},\dots,i_N,j_1,\dots,j_{m-1},j_{m+1},\dots,j_M}=\sum_{i_n=1}^{I_n}a_{i_1,\dots,i_{n-1},i_n,i_{n+1},\dots,i_N}b_{j_1,\dots,j_{m-1},i_n,j_{m+1},\dots,j_M}$. Tensor contractions are always differentiable and hence compatible with most automatic differentiation packages provided by popular deep learning libraries.

\subsubsection{Special Tensor Contractions}

A $(2, 1)$-tensor contraction between two order-$2$ tensors, $\textbf{A} \in \mathbb{R}^{I_1 \times I_2}$ and $\textbf{B} \in \mathbb{R}^{J_1 \times J_2}$, where $I_2 = J_1$, is equivalent to a standard matrix multiplication, $\textbf{C} = \textbf{A} \times_2^1 \textbf{B} = \textbf{A} \textbf{B} \in \mathbb{R}^{I_1 \times J_2}$. Similarly, a $(2, 1)$-tensor contraction between an order-$2$ tensor, $\textbf{A} \in \mathbb{R}^{I_1 \times I_2}$, and an order-$1$ tensor, $\textbf{b} \in \mathbb{R}^{J_1}$, where $I_2 = J_1$, is equivalent to a standard matrix-by-vector multiplication, $\textbf{c} = \textbf{A} \times_2^1 \textbf{b} = \textbf{A} \textbf{b} \in \mathbb{R}^{I_1}$.

\subsubsection{Mode-$n$ Product} \label{sec:mode_n_product}

\begin{figure}[h]
    \centering
    \includegraphics[width=1.0\linewidth]{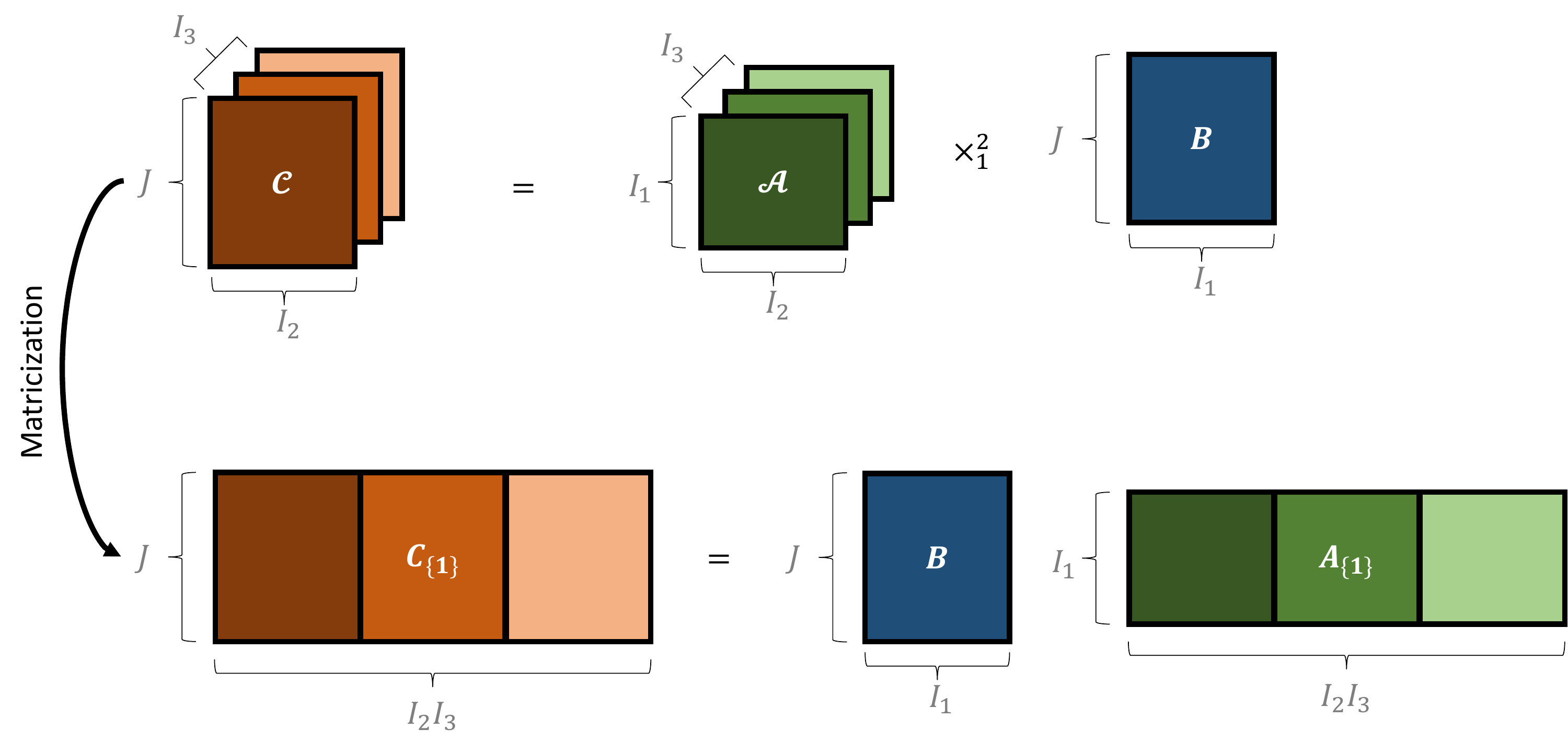}
    \caption{Tensor (top) and matricized (bottom) view of a mode-$n=1$ product between an order-$3$ tensor, $\mathcalbf{A} \in \mathbb{R}^{I_1 \times I_2 \times I_3}$, and matrix, $\textbf{B} \in \mathbb{R}^{J \times I_1}$, to yield, $\mathcalbf{C} \in \mathbb{R}^{J \times I_2 \times I_3}$.}
    \label{fig:mode_n_product}
\end{figure}

A mode-$n$ product, $\mathcalbf{C} = \mathcalbf{A} \times_n^2 \textbf{B}$, between an order-$N$ tensor, $\mathcalbf{A} \in \mathbb{R}^{I_1 \times \dots \times I_N}$, and a matrix, $\textbf{B} \in \mathbb{R}^{J \times I_n}$, yields another order-$N$ tensor, $\mathcalbf{C} \in \mathbb{R}^{I_1 \times I_{n-1} \times J \times I_{n+1} \times \dots \times J_N}$. This is equivalent to the matrix multiplication, $\textbf{C}_{ \{ n \}} = \textbf{B} \textbf{A}_{\{n\}}$, followed by a tensorization operation, as illustrated in Figure \ref{fig:mode_n_product}.

\subsubsection{Tucker Product} \label{sec:tucker_prod}

A Tucker product of the form, $\mathcalbf{C} = \llbracket \mathcalbf{A}; \textbf{B}^{(1)}, \ldots, \textbf{B}^{(N)} \rrbracket$, between an order-$N$ tensor, $\mathcalbf{A} \in \mathbb{R}^{I_1 \times \dots \times I_N}$, and $N$ matrices, $\textbf{B}^{(n)} \in \mathbb{R}^{J_n \times I_n}$, $n=1,\ldots,N$, yields another order-$N$ tensor, $\mathcalbf{C} \in \mathbb{R}^{J_1 \times \dots \times J_N}$, through $N$ mode-$n$ products, $\mathcalbf{C} = \mathcalbf{A} \times_1^2 \textbf{B}^{(1)} \times_2^2 \textbf{B}^{(2)} \times_3^2 \cdots \times_N^2 \textbf{B}^{(N)}$. The Tucker product can also be written in vectorized form as $\vectorize \left( \mathcalbf{C} \right) = \left( \bigotimes_{n=1}^N \textbf{B}^{(n)} \right) \vectorize \left( \mathcalbf{A} \right)$, where $\bigotimes_{n=1}^N \textbf{B}^{(n)} = \textbf{B}^{(1)} \otimes \cdots \otimes \textbf{B}^{(N)}$.


\subsubsection{Tensor-Train Decomposition (TTD)} \label{sec:ttd}

The Matrix-Product Operator (MPO) form of the Tensor-Train Decomposition (TTD) approximates a large order-$2N$ tensor, $\mathcalbf{A} \in \mathbb{R}^{(I_1 \times J_1) \times \dots \times (I_N \times J_N)}$, via $N$ smaller core tensors, $\mathcalbf{G}^{(n)} \in \mathbb{R}^{R_{n-1} \times I_n \times J_n \times R_n}$, as $\mathcalbf{A} = \mathcalbf{G}^{(1)} \times^1_4 \mathcalbf{G}^{(2)} \times^1_4 \cdots \times^1_4 \mathcalbf{G}^{(N)}$. The set $\{ R_0, \ldots, R_N \}$, where $R_0=R_N=1$, is referred to as the TT-rank. The TTD reduces the space complexity of the orignal tensor from an exponential $\prod_{n=1}^{N} I_n J_n$ to a linear $\sum_{n=1}^N I_n J_n R_{n-1} R_n$ in the dimensions $I_n$ and $J_n$, which is highly efficient for small values of $I_n$, $J_n$, and $R_n$. For $I_n=J_n=R_n=2$, TTD achieves the optimal compression ratio and is referred to as the Quantized TTD (QTTD). 

\subsubsection{Convolution Tensor} \label{sec:convolution_tensor}

The Convolution tensor, $\mathcalbf{C} \in \mathbb{R}^{I \times I \times P}$, where $I>P$, is a sparse order-$3$ tensor with entries $t_{i, (i+p-1)\%I, p} = 1$ for $i=1,\ldots,I$ and $p=1,\ldots,P$, where $\%$ denotes the modulo operation. Given an input vector, $\textbf{x} \in \mathbb{R}^{I}$, and a convolution kernel, $\textbf{k} \in \mathbb{R}^{P}$, the discrete convolution can be obtained through the tensor contraction $\textbf{y} = \textbf{x} \circledast \textbf{k} = \mathcalbf{C} \times_2^1 \textbf{x} \times_3^1 \textbf{k} = \textbf{S} \textbf{x} \in \mathbb{R}^{L}$, where $\textbf{S} = \mathcalbf{C} \times_3^1 \textbf{k} \in \mathbb{R}^{I \times I}$ is a circulant matrix generated from the convolution kernel, $\textbf{k}$. The circulant matrix, $\textbf{S}$, can be interpreted as the adjacency matrix of a circulant graph, which effectively casts the standard convolution as a special case of graph convolution on a circulant graph. The convolution tensor has an exact QTTD low-rank format.

\subsubsection{Data Tensor} \label{sec:data_tensor}

An order-$(N+M)$ data tensor, $\mathcalbf{X} \in \mathbb{R}^{I_1 \times \cdots \times I_N \times J_1 \times \cdots \times J_M}$, is a collection of input data over $N$ domain modes and $M$ feature modes. A domain mode of size $I_n$ is a mode with well-defined structure that can be described as a graph (e.g., the image pixels arranged as a regular grid graph), while a feature mode of size $J_m$ is a dimension with a collection of descriptive features (e.g., the RGB value of an image pixel). Such a data tensor is general enough to describe the inputs to most existing neural network architectures.


\subsubsection{Tensor Network Diagrams}

\begin{figure}[t]
    \centering
    \includegraphics[width=0.92\linewidth]{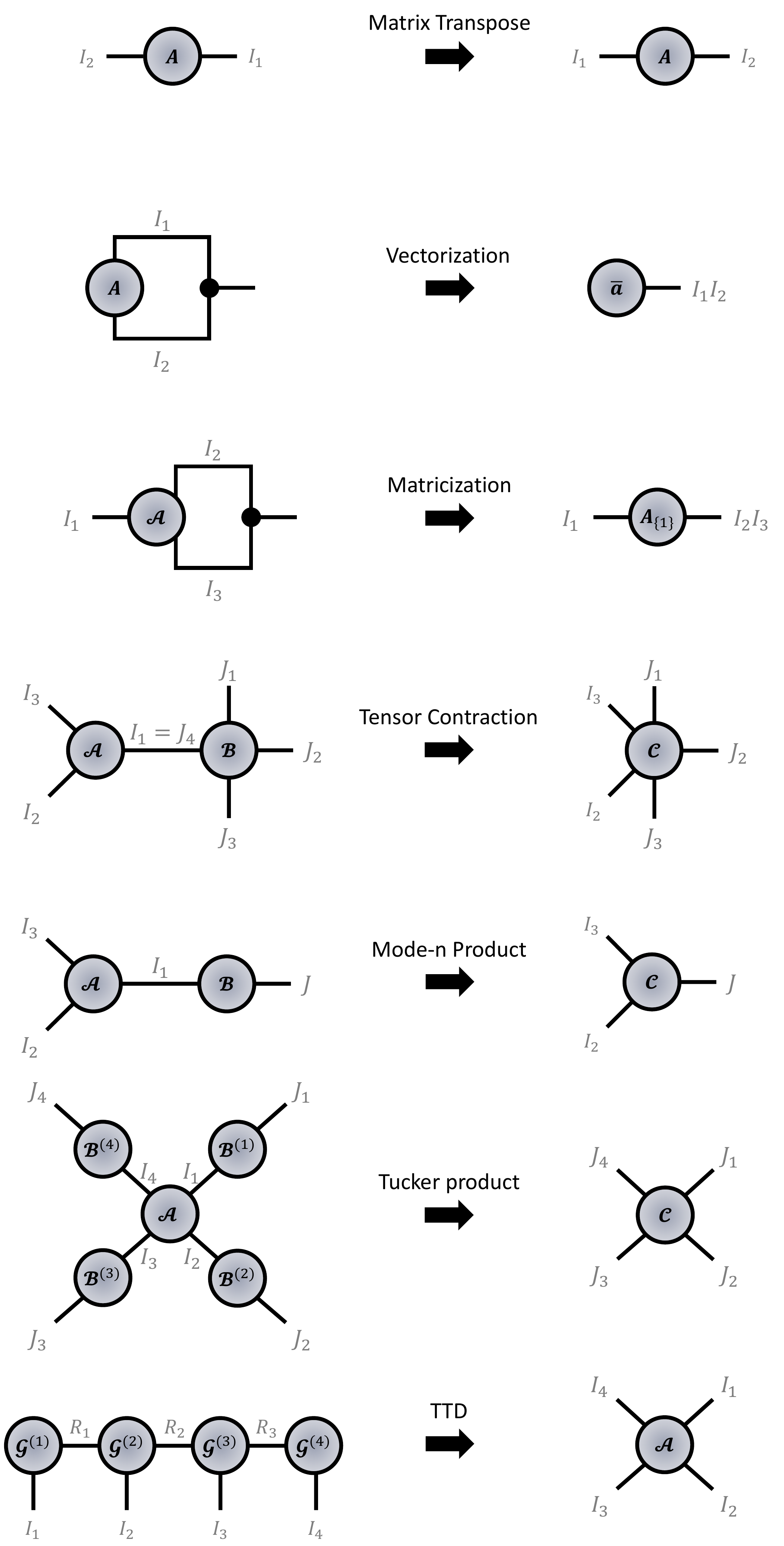}
    \caption{Tensor Network (TN) diagrams for basic tensor operations: 
    (1) Matrix transpose from $\mathbf{A} \in \mathbb{R}^{I_1 \times I_2}$ to $\mathbf{A}^T \in \mathbb{R}^{I_2 \times I_1}$; 
    (2) Matrix vectorization from $\mathbf{A} \in \mathbb{R}^{I_1 \times I_2}$ to $\mathbf{\bar{a}} \in \mathbb{R}^{I_1 I_2}$;
    (3) Tensor matricization from $\mathcalbf{A} \in \mathbb{R}^{I_1 \times I_2 \times I_3}$ to $\mathbf{A}_{\{ 1 \}} \in \mathbb{R}^{I_1 \times I_2 I_3}$; 
    (4) Tensor contraction, $\mathcalbf{C} = \mathcalbf{A} \times_1^4 \mathcalbf{B}$, with $\mathcalbf{A} \in \mathbb{R}^{I_1 \times I_2 \times I_3}$ and $\mathcalbf{B} \in \mathbb{R}^{J_1 \times J_2 \times J_3 \times J_4}$ to yield $\mathcalbf{C} \in \mathbb{R}^{I_2 \times I_3 \times J_1 \times J_2 \times J_3}$; 
    (5) Mode-$n$ product, $\mathcalbf{C} = \mathcalbf{A} \times_2^2 \mathbf{B}$, with $\mathcalbf{A} \in \mathbb{R}^{I_1 \times I_2 \times I_3}$ and $\mathbf{B} \in \mathbb{R}^{I_1 \times J}$ to yield $\mathcalbf{C} \in \mathbb{R}^{I_2 \times I_3 \times J}$;
    (6) Tucker product, $\mathcalbf{C} = \llbracket \mathcalbf{A}; \textbf{B}^{(1)}, \textbf{B}^{(2)}, \textbf{B}^{(3)}, \textbf{B}^{(4)} \rrbracket$, with $\mathcalbf{A} \in \mathbb{R}^{I_1 \times I_2 \times I_3 \times I_4}$ and $\textbf{B}^{(n)} \in \mathbb{R}^{I_n \times J_n}$ for $n=1,\ldots,4$ to yield $\mathcalbf{C} \in \mathbb{R}^{J_1 \times J_2 \times J_3 \times J_4}$; 
    (7) Tensor-Train Decomposition (TTD), $\mathcalbf{A} = \mathcalbf{G}^{(1)} \times^1_4 \mathcalbf{G}^{(2)} \times^1_4 \mathcalbf{G}^{(3)} \times^1_4 \mathcalbf{G}^{(4)}$, with $\mathcalbf{G}^{(n)} \in \mathbb{R}^{R_{n-1} \times I_n \times J_n \times R_n}$ for $n=1,\ldots,4$ to yield $\mathcalbf{A} \in \mathbb{R}^{I_1 \times I_2 \times I_3 \times I_4}$.
    }
    \label{fig:basic_tn}
\end{figure}

A Tensor Network (TN) diagram is an intuitive yet mathematically rigorous way of visualizing tensor operations. In a TN diagram, a vertex with $N$ open edges represents an order-$N$ tensor, while a pair of vertices connected through a common edge represents a tensor contraction between two tensors over their modes of common size. Basic tensor operations in the TN notation are illustrated in Figure \ref{fig:basic_tn}.

\subsection{Basic Graph Signal Processing}

\subsubsection{Definitions} 

A graph is defined by a set of $I$ vertices (or nodes) and a set of edges connecting pairs of vertices. A graph can be described by its Graph Shift Operator (GSO), $\textbf{S} \in \mathbb{R}^{I_1 \times I_2}$, $I_1=I_2=I$. Special cases of GSOs include the weighted graph adjacency matrix, where the entry of the GSO (the edge weight), $s_{i_1, i_2} \in \mathbb{R}$, is positive if there exists an edge that connects the $i_1$-th and $i_2$-th vertices, and zero otherwise. 



\subsubsection{Graph Inference} \label{sec:graph_inference}

Given a set of vertices and their descriptive features, an unknown graph topology can be inferred through the use of a pair-wise similarity function, $f(\cdot)$, such that the edge weights are estimated as $s_{i_1, i_2} = f \left( \textbf{x}^{(i_1)}, \textbf{x}^{(i_2)} \right)$, where $\textbf{x}^{(i_1)}$ and $\textbf{x}^{(i_2)}$ are the features associated with the $i_1$-th and $i_2$-th vertices. 

\subsubsection{Graph Signals}

A graph signal, $\textbf{x} \in \mathbb{R}^{I}$, is a vector of size $I$ that associates a scalar value to each of the $I$ vertices in the graph (e.g., the age of users in a social network graph). Given a set of $J$ graph signals, these can be stacked as column vectors to form the graph signal matrix, $\textbf{X} \in \mathbb{R}^{I \times J}$.

\subsubsection{Graph Signal Processing (GSP) via GSOs} \label{sec:prelim_gso}

The matrix-by-vector product between a GSO, $\textbf{S} \in \mathbb{R}^{I \times I}$, and a graph signal, $\textbf{x} \in \mathbb{R}^{I}$, generates another graph signal,  $\textbf{y} = \textbf{S} \textbf{x} \in \mathbb{R}^{I}$, which can have several well-defined physical meanings depending on the choice of the GSO \cite{scalzo2023class}. For instance, a graph adjacency matrix based GSO performs a neighbourhood aggregation operation where the graph signal at the $i_1$-th vertex, $y_{i_1}$, results from the sum of graph signals from its neighbouring (i.e., connected) vertices, $y_{i_1} = \sum_{i_2} s_{i_1. i_2} x_{i_2}$. This can be naturally extended to graph signal matrices for $J$ graph signals as $\textbf{Y} = \textbf{S} \textbf{X} \in \mathbb{R}^{I \times J}$. 

\section{The Graph Tensor Network Framework} \label{sec:gtn}

Given the ability of graphs to operate on both irregular and regular domains, the ability of tensors to manipulate data across several dimensions, and the universal function approximation property of neural networks, it is therefore natural to investigate their conjoint treatment. To this end, we introduce the Graph Tensor Network (GTN) framework, which is suitable for both describing and designing neural networks that handles large-dimensional data on regular and irregular domains. The GTN framework is next shown to be general enough to include most modern neural network architectures as special cases. By virtue of graphs and tensors, the proposed GTN is flexible enough to handle tensor-variate data on any and many domains. In addition, through the use of Tensor Network (TN) diagrams, the proposed framework allows for the design of large-scale neural learning systems in a highly intuitive and mathematically rigorous manner.

\subsection{The Forward Pass}

\begin{figure}[h]
    \centering
    \includegraphics[width=1.0\linewidth]{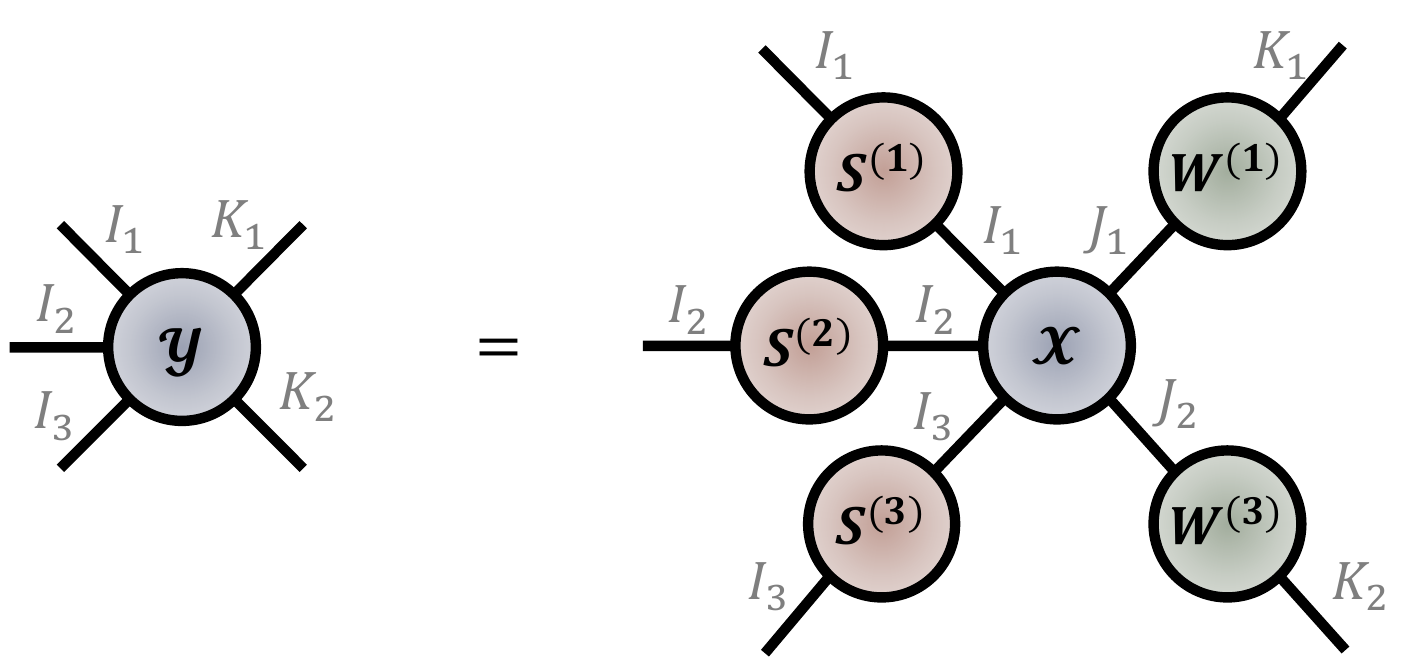}
    \caption{The forward pass of a Graph Tensor Network (GTN) for an order-$(3+2)$ input data tensor, $\mathcalbf{X} \in \mathbb{R}^{I_1 \times I_2 \times I_3 \times J_1 \times J_2}$, according to (\ref{eq:gtn_general}).}
    \label{fig:gtn_general}
\end{figure}

Consider a machine learning paradigm where the input tensor, $\mathcalbf{X} \in \mathbb{R}^{I_1 \times \cdots \times I_N \times J_1 \times \cdots \times J_M}$, is an order-$\left( N+M \right)$ data tensor with $N$ domain modes and $M$ feature modes, as defined in Section \ref{sec:data_tensor}. The proposed Graph Tensor Network (GTN) framework performs a Tucker product forward pass, given by
\begin{equation} \label{eq:gtn_general}
    \begin{aligned}
        \mathcalbf{Y} = \llbracket \mathcalbf{X}; \textbf{S}^{(1)}, \ldots, \textbf{S}^{(N)}, \textbf{W}^{(1)}, \ldots, \textbf{W}^{(M)} \rrbracket 
    \end{aligned}
\end{equation}
where $\textbf{S}^{(n)} \in \mathbb{R}^{I_n \times I_n}$ is the mode-$n$ Graph Shift Operator (GSO) applied to the $n$-th domain mode (there are as many graph filters as the number of domain modes) and $\textbf{W}^{(m)} \in \mathbb{R}^{K_m \times J_m}$ is the mode-$m$ weight matrix applied to the $m$-th feature mode (there are as many weight matrices as the number of feature modes). Figure \ref{fig:gtn_general} illustrates the forward pass in (\ref{eq:gtn_general}) in the Tensor Network (TN) notation. Similar to classical neural networks, a bias tensor, $\mathcalbf{B} \in \mathbb{R}^{I_1 \times \cdots \times I_N \times J_1 \times \cdots \times J_M}$, can be optionally added to the final transform, followed by an activation function to introduce non-linearity. 

\begin{remark}
The dimensions of the tensor operations are unchanged by the application of an activation function due to its element-wise nature. This implies that the TN topology is independent of the activation function.
\end{remark}

\begin{remark}
The mode-$n$ products between the input tensor, $\mathcalbf{X} \in \mathbb{R}^{I_1 \times \cdots \times I_N \times J_1 \times \cdots \times J_M}$, and $N$ GSOs, $\textbf{S}^{(n)} \in \mathbb{R}^{I_n \times I_n}$, can be interpreted as $N$ simultaneous graph shift operations over all $N$ domain modes of the input tensor. Indeed, each mode-$n$ product, $\mathcalbf{X} \times_n^2 \textbf{S}^{(n)}$, can be interpreted as a mode-$n$ graph operation over the graph signal matrix, $\mathbf{X}_{\{n\}} \in \mathbb{R}^{I_n \times I_1 \cdots I_{n-1} I_{n+1} \cdots I_N}$, as $\textbf{Y} = \textbf{S}^{(n)} \textbf{X}_{\{ n \}}$, which follows from the property discussed in Section \ref{sec:mode_n_product}. 
\end{remark}



\section{Classical Neural Networks as GTNs} \label{sec:classical_nnas}

This section derives several neural network architectures as special cases of the GTN forward pass in (\ref{eq:gtn_general}), whereby the difference in architectures is primarily captured through the design of the corresponding GSO for the domain mode. 

\begin{remark}
For conciseness, the derivations in this section will omit the bias term and the activation function. However, these can always be added to the final transform without any loss of generality.
\end{remark}

\subsection{Dense Neural Network}

\begin{figure}[h]
    \centering
    \includegraphics[width=0.7\linewidth]{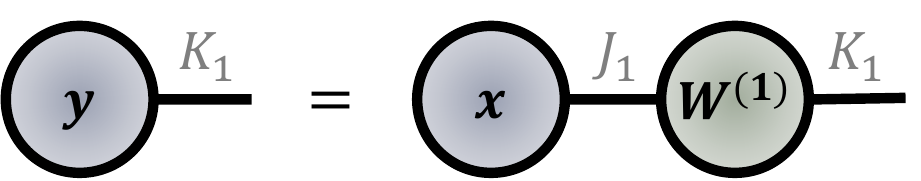}
    \caption{Dense Neural Network as a Graph Tensor Network with $N=0$ domain modes and $M=1$ feature mode, according to (\ref{eq:gtn_dnn}).}
    \label{fig:gtn_dnn}
\end{figure}

A Dense Neural Network (DNN) takes as input an order-$(0+1)$ data tensor as defined in Section \ref{sec:data_tensor} (a feature vector with $N=0$ domain mode and $M=1$ feature mode), $\textbf{x} \in \mathbb{R}^{J_1}$, and generates an output vector according to the transform
\begin{equation} \label{eq:dnn_full}
    \textbf{y} = \textbf{W} \textbf{x} \in \mathbb{R}^{K_1}
\end{equation}
where $\textbf{W} \in \mathbb{R}^{K_1 \times J_1}$ is a trainable weight matrix. The DNN forward pass in (\ref{eq:dnn_full}) can be shown to be a special case of (\ref{eq:gtn_general}) with $N=0$ domain modes and $M=1$ feature mode as
\begin{equation} \label{eq:gtn_dnn}
    \begin{aligned}
        \mathbf{y} 
        &= \mathbf{W} \mathbf{x} \\
        &= \mathbf{x} \times_{1}^2 \mathbf{W} \\
        &= \llbracket \mathbf{x}; \mathbf{W} \rrbracket \\
        &= \llbracket \mathbf{x}; \mathbf{W}^{(1)} \rrbracket \\
    \end{aligned}
\end{equation}
where ${\textbf{W}^{(1)}} = \textbf{W}$ for simplicity of notation. Figure \ref{fig:gtn_dnn} illustrates (\ref{eq:gtn_dnn}) in the TN notation.

\subsection{Graph Convolutional Network} \label{sec:gtn_gnn}

\begin{figure}[h]
    \centering
    \includegraphics[width=1.0\linewidth]{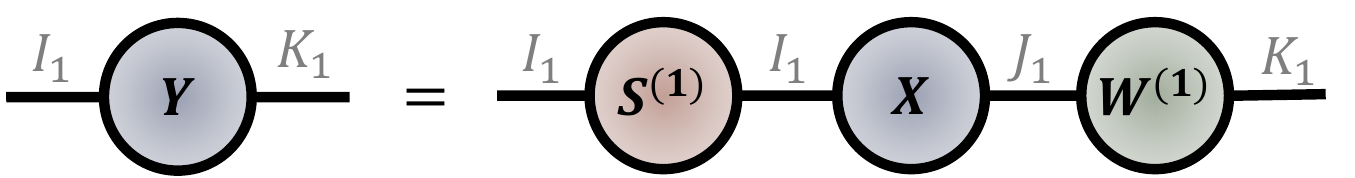}
    \caption{Graph Convolutional Network as a Graph Tensor Network with $N=1$ domain mode (representing graph vertices) and $M=1$ feature mode, according to (\ref{eq:gtn_gnn}).}
    \label{fig:gtn_gnn}
\end{figure}

Given the adjacency matrix of a graph, $\textbf{A} \in \mathbb{R}^{I_1 \times I_1}$, a Graph Convolutional Network (GCN) \cite{kipf2016semi} takes as input an order-$(1+1)$ data tensor (a graph signal matrix with $N=1$ graph domain mode and $M=1$ feature mode), $\textbf{X} \in \mathbb{R}^{I_1 \times J_1}$, and generates an output matrix according to the transform
\begin{equation} \label{eq:gnn_full}
    \textbf{Y} = \textbf{S} \textbf{X} \textbf{W} \in \mathbb{R}^{I_1 \times K_1}
\end{equation}
where $\textbf{W} \in \mathbb{R}^{J_1 \times K_1}$ is a trainable weight matrix and $\textbf{S} = \tilde{\textbf{D}}^{-\frac{1}{2}} \tilde{\textbf{A}} \tilde{\textbf{D}}^{-\frac{1}{2}} \in \mathbb{R}^{I_1 \times I_1}$ is a graph convolution operator with $\tilde{\mathbf{A}} = \textbf{I} + \textbf{A} \in \mathbb{R}^{I_1 \times I_1}$ and $\tilde{\mathbf{D}} = \degreemat \left( \tilde{\mathbf{A}} \right)$. The GCN forward pass in (\ref{eq:gnn_full}) can be shown to be a special case of (\ref{eq:gtn_general}) with $N=1$ domain mode (representing graph vertices) and $M=1$ feature mode as 
\begin{equation} \label{eq:gtn_gnn}
    \begin{aligned}
        \mathbf{Y} 
        &= \textbf{S} \textbf{X} \textbf{W} \\
        &= \mathbf{X} \times_1^2 \mathbf{S} \times_{2}^2 \mathbf{W}^T \\
        &= \llbracket \mathbf{X}; \mathbf{S}, \mathbf{W}^T \rrbracket \\
        &= \llbracket \mathbf{X}; \mathbf{S}^{(1)}, \mathbf{W}^{(1)} \rrbracket \\
    \end{aligned}
\end{equation}
where $\textbf{S} = {\textbf{S}^{(1)}}$ and $\textbf{W}^{(1)} = {\textbf{W}}^T$ for simplicity of notation. Figure \ref{fig:gtn_gnn} illustrates (\ref{eq:gtn_gnn}) in the TN notation.

\subsection{Convolutional Neural Network} \label{sec:gtn_cnn}

\begin{figure}[h]
    \centering
    \includegraphics[width=0.7\linewidth]{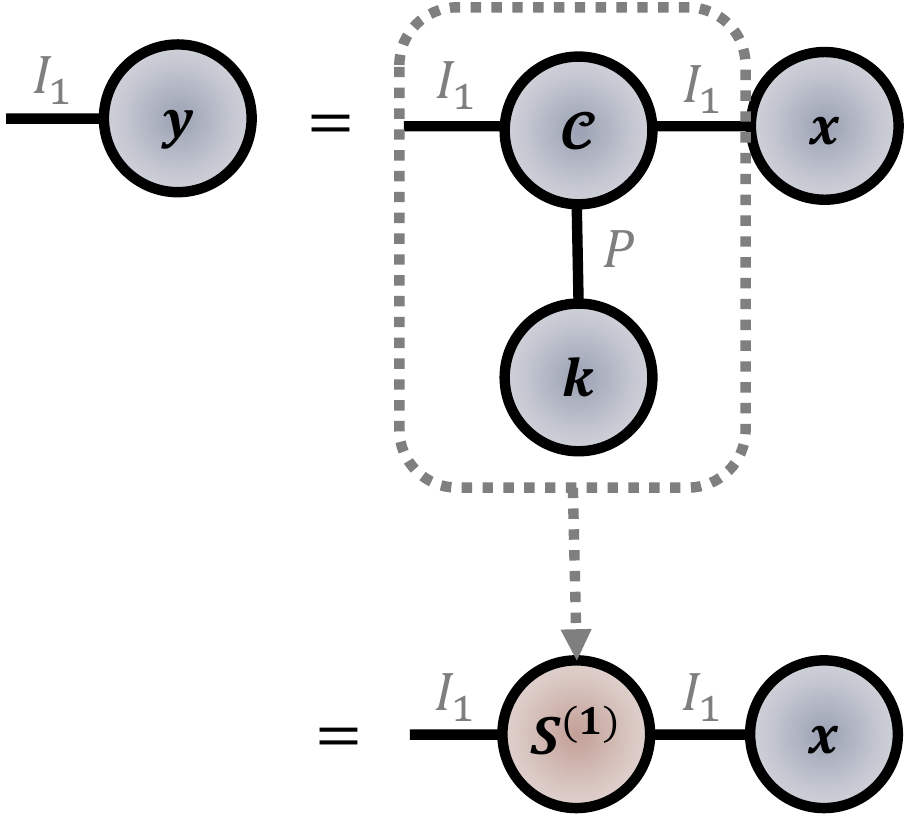}
    \caption{Convolutional Neural Network as a Graph Tensor Network with $N=1$ domain mode and $M=0$ feature mode, according to (\ref{eq:gtn_cnn}).}
    \label{fig:gtn_cnn}
\end{figure}

A Convolutional Neural Network (CNN) takes as input an order-$(1 + 0)$ data tensor (where the $N=1$ domain mode corresponds to the convolution domain), $\mathbf{x} \in \mathbb{R}^{I_1}$, and generates the output vector through the discrete convolution
\begin{equation} \label{eq:cnn_full}
    \mathbf{y} = \mathbf{x} \circledast \mathbf{k} \in \mathbb{R}^{I_1}
\end{equation}
where $\mathbf{k} \in \mathbb{R}^{P}$ is a trainable convolution kernel. By applying the convolution tensor defined in Section \ref{sec:convolution_tensor}, $\mathcalbf{C} \in \mathbb{R}^{I \times I \times P}$, the CNN forward pass in (\ref{eq:cnn_full}) can be shown to be a special case of (\ref{eq:gtn_general}) with $N=1$ domain mode (representing the convolution domain) and $M=0$ feature mode as 
\begin{equation} \label{eq:gtn_cnn}
    \begin{aligned}
        \mathbf{y} 
        &= \textbf{x} \circledast \textbf{k}\\
        &= \mathcalbf{C} \times_2^1 \textbf{x} \times_3^1 \textbf{k} \\  
        &= \left( \mathcalbf{C} \times_3^1 \textbf{k} \right) \textbf{x} \\  
        &= \textbf{S}^{(1)} \mathbf{x} \\
        &= \mathbf{x} \times_{1}^2 \textbf{S}^{(1)} \\        
        &= \llbracket \mathbf{x}; \mathbf{S}^{(1)} \rrbracket \\
    \end{aligned}
\end{equation}
where ${\textbf{S}^{(1)}} = \mathcalbf{C} \times_3^1 \textbf{k}$. Figure \ref{fig:gtn_cnn} illustrates (\ref{eq:gtn_cnn}) in the TN notation.

\begin{remark}
The GSO, ${\textbf{S}^{(1)}} = \mathcalbf{C} \times_3^1 \textbf{k}$, is a circulant matrix constructed from the trainable convolution kernel, $\textbf{k} \in \mathbb{R}^P$, which effectively acts as the adjacency matrix of a circulant graph, as discussed in Section \ref{sec:convolution_tensor}.
\end{remark}

\subsection{Attention Neural Network}

\begin{figure} [h]
    \centering
    \includegraphics[width=1.0\linewidth]{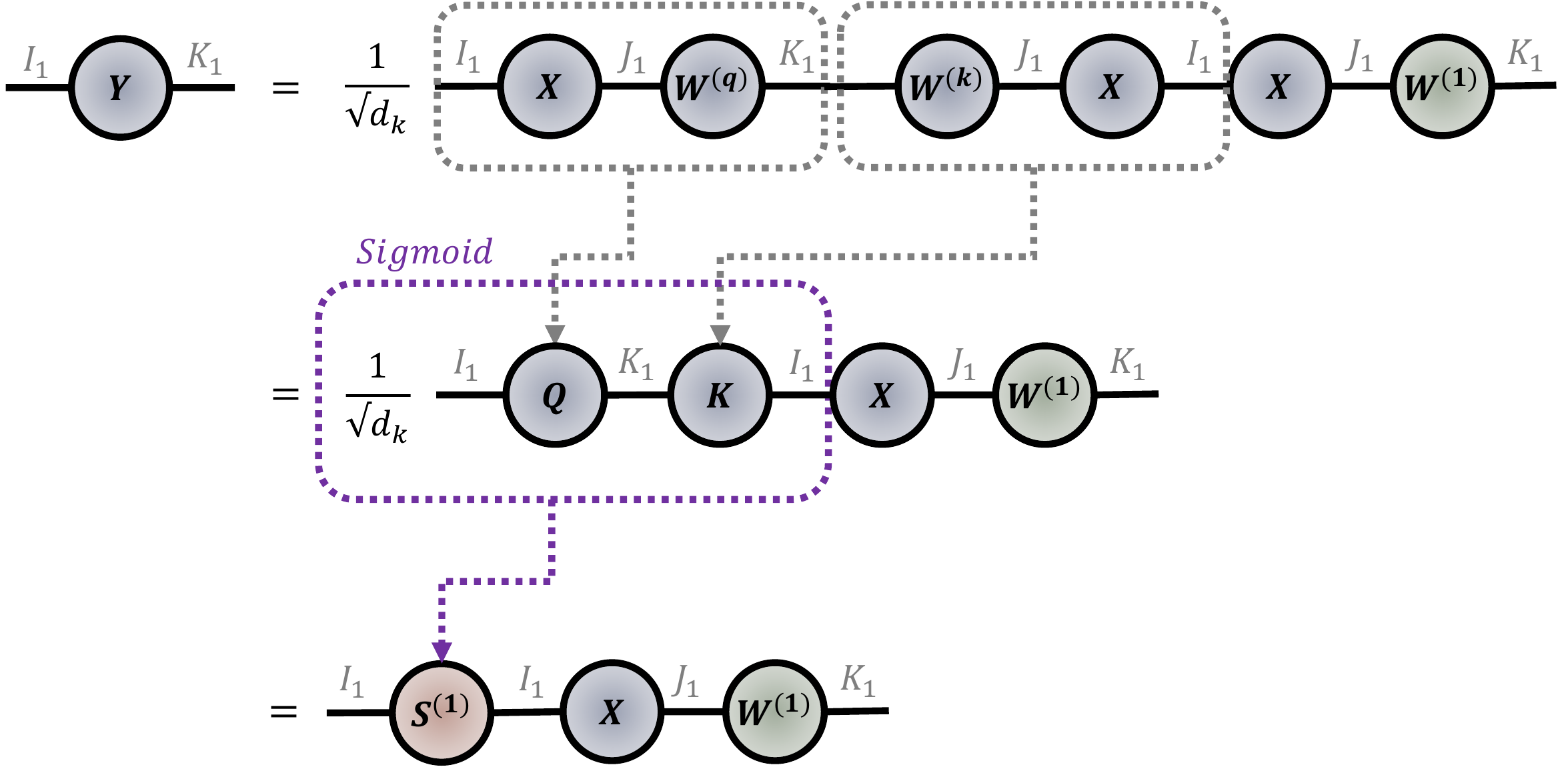}
    \caption{Attention Neural Network as a Graph Tensor Network with $N=1$ domain mode (representing time) and $M=1$ feature mode, according to (\ref{eq:gtn_ann}).}
    \label{fig:gtn_ann}
\end{figure}

A dot-product style Attention Neural Network (ANN) \cite{vaswani2017attention}, takes as input an order-$(1+1)$ data tensor (a time-series features matrix with $N=1$ time-domain mode and $M=1$ feature mode), $\textbf{X} \in \mathbb{R}^{I_1 \times J_1}$, and generates the output matrix according to the transform
\begin{equation} \label{eq:ann_full}
    \begin{aligned}
    \textbf{Y}
    &= \sigma \left( \frac{1}{\sqrt{d_k}} \left( \textbf{X} \textbf{W}^{(q)} \right) \left( \textbf{X} \textbf{W}^{(k)} \right)^T \right) \textbf{X} \textbf{W}^{(v)} \\
    \end{aligned}
\end{equation}
where $\textbf{W}^{(k)} \in \mathbb{R}^{J_1 \times K_1}$, $\textbf{W}^{(q)} \in \mathbb{R}^{J_1 \times K_1}$, and $\textbf{W}^{(v)} \in \mathbb{R}^{J_1 \times K_1}$ are trainable weight matrices, $\sigma(\cdot)$ is a $SoftMax$ activation function, and $\sqrt{d_k}$ is a scaling factor. The attention operation can also be written in a short-hand form as $\textbf{Y} = \sigma \left( \frac{1}{\sqrt{d_k}} \textbf{Q}\textbf{K}^T \right) \textbf{V}$, where $\textbf{Q} = \textbf{X} \textbf{W}^{(q)} \in \mathbb{R}^{I_1 \times K_1}$, $\textbf{K} = \textbf{X} \textbf{W}^{(k)} \in \mathbb{R}^{I_1 \times K_1}$, and $\textbf{V} = \textbf{X} \textbf{W}^{(v)} \in \mathbb{R}^{I_1 \times K_1}$ are referred to respectively as the Key, Query, and Value matrices. The ANN forward pass in (\ref{eq:ann_full}) can be shown to be a special case of (\ref{eq:gtn_general}) with $N=1$ domain mode (representing time) and $M=1$ feature mode, to yield
\begin{equation} \label{eq:gtn_ann}
    \begin{aligned}
        \mathbf{Y} 
        &= \sigma \left( \frac{1}{\sqrt{d_k}} \left( \textbf{X} \textbf{W}^{(q)} \right) \left( \textbf{X} \textbf{W}^{(k)} \right)^T \right) \textbf{X} \textbf{W}^{(v)} \\
        &= \sigma \left( \frac{1}{\sqrt{d_k}} \textbf{Q} \textbf{K}^T \right) \textbf{X} \textbf{W}^{(v)} \\
        &= \textbf{S}^{(1)} \textbf{X} \textbf{W}^{(v)} \\
        &= \mathbf{X} \times_1^2 \mathbf{S}^{(1)} \times_{2}^2 \mathbf{W}^{(v)} \\  
        &= \llbracket \mathbf{X}; \mathbf{S}^{(1)}, \mathbf{W}^{(v)} \rrbracket \\
        &= \llbracket \mathbf{X}; \mathbf{S}^{(1)}, \mathbf{W}^{(1)} \rrbracket \\    
    \end{aligned}
\end{equation}
where $\mathbf{S}^{(1)} = \sigma \left( \frac{1}{\sqrt{d_k}} \textbf{Q} \textbf{K}^T \right)$ and $\textbf{W}^{(1)} = {\mathbf{W}^{(v)}}$ for simplicity of notation. Figure \ref{fig:gtn_ann} illustrates (\ref{eq:gtn_ann}) in the TN notation.

\begin{remark}
The GSO, $\mathbf{S}^{(1)} = \sigma \left( \frac{1}{\sqrt{d_k}} \textbf{Q} \textbf{K}^T \right)$, where $\textbf{Q} = \textbf{X} \textbf{W}^{(q)} \in \mathbb{R}^{I_1 \times K_1}$ and $\textbf{K} = \textbf{X} \textbf{W}^{(k)} \in \mathbb{R}^{I_1 \times K_1}$, can be interpreted as the adjacency matrix of a time-domain graph, where the vertices represent different time-steps. In this graph, the edge weight connecting pairs of time-steps is inferred through a pair-wise similarity function parameterized by trainable weight matrices $\textbf{W}^{(q)}$ and $\textbf{W}^{(k)}$. This allows the attention mechanism to construct a different graph topology depending on the input itself, which is the core of the self-attention mechanism. 
\end{remark}

\subsection{Recurrent Neural Network} \label{sec:rnn_as_gtn}

\begin{figure}[h]
    \centering
    \includegraphics[width=1.0\linewidth]{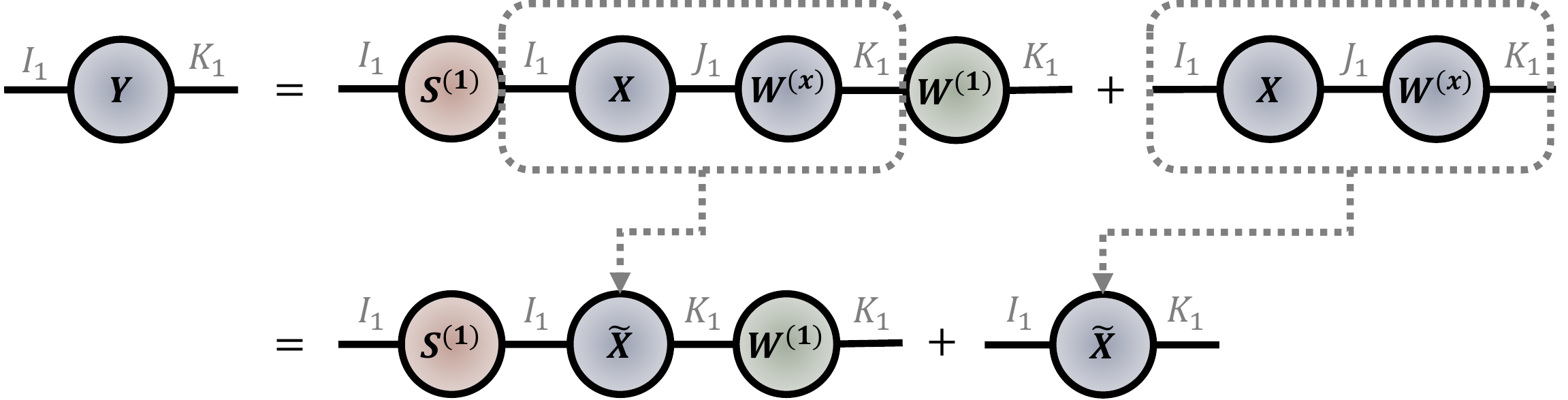}
    \caption{Recurrent Neural Network as a Graph Tensor Network with $N=1$ domain mode (representing time) and $M=1$ feature mode, according to (\ref{eq:derivation_gtn_rnn}).}
    \label{fig:gtn_rnn}
\end{figure}

A Recurrent Neural Network (RNN) \cite{mandic2001recurrent} takes as input an order-$(1+1)$ data tensor (a time-series features matrix with $N=1$ time-domain mode and $M=1$ feature mode), $\textbf{X} \in \mathbb{R}^{I_1 \times J_1}$, where the $i_1$-th row vector, $\textbf{x}_{i_1} \in \mathbb{R}^{J_1}$, is the feature vector at the $i_1$-th time-step. The RNN is characterized by the recurrent computation of hidden states
\begin{equation} \label{eq:rnn_full}
\textbf{y}_{i_1} = \textbf{W}^{(r)} \textbf{y}_{i_1-1} + \textbf{W}^{(x)} \textbf{x}_{i_1}    
\end{equation}
where $\textbf{W}^{(r)} \in \mathbb{R}^{K_1 \times K_1}$ is a trainable recurrent weight matrix, $\textbf{y}_{i_1-1} \in \mathbb{R}^{K_1}$ is the hidden state from the previous time-step, $\textbf{W}^{(x)}  \in \mathbb{R}^{K_1 \times J_1}$ is a trainable input weight matrix, and $\textbf{x}_{i_1} \in \mathbb{R}^{J_1}$ is the input feature vector at the current time-step. The hidden state equation in (\ref{eq:rnn_full}) can be shown to be a special case of (\ref{eq:gtn_general}) with $N=1$ domain mode (representing time) and $M=1$ feature mode. This is achieved by expanding the recurrent computation as 
\begin{equation} \label{eq:rnn_hidden_states_derivation}
    \begin{aligned}
        \textbf{y}_{i_1} 
        &= {\textbf{W}^{(r)}} \textbf{y}_{i_1-1} + {\textbf{W}^{(x)}} {\textbf{x}}_{i_1} \\
        &= {\textbf{W}^{(r)}} \textbf{y}_{i_1-1} + \tilde{\textbf{x}}_{i_1} \\
        &= {\textbf{W}^{(r)}} \left( {\textbf{W}^{(r)}}^{i_1-2} \tilde{\textbf{x}}_{1} + \cdots + \tilde{\textbf{x}}_{i_1-1} \right) + \tilde{\textbf{x}}_{i_1} \\        
        &= {\textbf{W}^{(r)}}^{i_1-1} \tilde{\textbf{x}}_1 + {\textbf{W}^{(r)}}^{i_1-2} \tilde{\textbf{x}}_2 + \cdots + \tilde{\textbf{x}}_{i_1}  \\
    \end{aligned}
\end{equation}
where $\tilde{\textbf{x}}_{i_1} = {\textbf{W}^{(x)}} {\textbf{x}}_{i_1}$ for simplicity of notation. This can then be written in a block-matrix form for $I_1$ time-steps as
\begin{equation} \label{eq:psi_vis_3}
    \begin{bmatrix}
    \mathbf{y}_1 \\
    \mathbf{y}_2 \\
    \vdots \\
    \mathbf{y}_{I_1} \\
    \end{bmatrix}
     = 
    \begin{bmatrix}
    \textbf{I} & \textbf{0} & \cdots & \textbf{0} \\
    {\textbf{W}^{(r)}}^1 & \textbf{I} & \cdots & \textbf{0} \\
    \vdots & \vdots & \ddots & \vdots \\
    {\textbf{W}^{(r)}}^{I_1-1} & {\textbf{W}^{(r)}}^{I_1-2} & \cdots & \textbf{I}\\
    \end{bmatrix}
    \begin{bmatrix}
    \tilde{\textbf{x}}_1 \\
    \tilde{\textbf{x}}_2 \\
    \vdots \\
    \tilde{\textbf{x}}_{I_1} \\
    \end{bmatrix}
\end{equation}
By allowing the recurrent weight matrix to be a scaled idempotent matrix, $\textbf{W}^{(r)} = c \textbf{W}^{(1)}$, such that ${\textbf{W}^{(r)}}^N = c^N \textbf{W}^{(1)}$, (\ref{eq:psi_vis_3}) can be simplified into  
\begin{equation} \label{eq:psi_vis_1}
\begin{aligned}
    \begin{bmatrix}
    \mathbf{y}_1 \\
    \mathbf{y}_{2} \\
    \vdots \\
    \mathbf{y}_{I_1} \\
    \end{bmatrix}
    &= 
    \begin{bmatrix}
    \textbf{I} & \textbf{0} & \cdots & \textbf{0} \\
    c^{ 1 } {\mathbf{W}} & \textbf{I} & \cdots & \textbf{0}\\
    \vdots & \vdots & \ddots & \vdots \\
    c^{ I_1-1 } {\mathbf{W}} & c^{ I_1 - 2 } \mathbf{W} & \cdots & \textbf{I}\\
    \end{bmatrix}
    \begin{bmatrix}
    \tilde{\textbf{x}}_1 \\
    \tilde{\textbf{x}}_2 \\
    \vdots \\
    \tilde{\textbf{x}}_{I_1} \\
    \end{bmatrix} 
    \\
\end{aligned}
\end{equation}
This expression can be written compactly through the Kronecker product as $\vectorize \left( \textbf{Y} \right) = \left( \textbf{S}^{(1)} \otimes \textbf{W}^{(1)} + \textbf{I} \right) \vectorize \left( \tilde{\mathbf{X}} \right)$, where $s^{(1)}_{i_1, i_2} = c^{ i_1 - i_2 }$ for $i_1 > i_2 $ and $s^{(1)}_{i_1, i_2}=0$ otherwise. By using the vectorized Tucker product property in Section \ref{sec:tucker_prod}, the matricization of (\ref{eq:psi_vis_1}) can be written as
\begin{equation} \label{eq:derivation_gtn_rnn}
    \begin{aligned}
        \textbf{Y}
        &= \matricize \left( \left( \textbf{S}^{(1)} \otimes \textbf{W}^{(1)} + \textbf{I} \right) \vectorize \left( \tilde{\mathbf{X}} \right) \right) \\
        &= \matricize \left( \left( \textbf{S}^{(1)} \otimes \textbf{W}^{(1)} \right) \vectorize \left( \tilde{\mathbf{X}} \right) + \vectorize \left( \tilde{\mathbf{X}} \right) \right) \\
        &= \matricize \left( \vectorize \left( \llbracket \tilde{\mathbf{X}}; \mathbf{S}^{(1)}, \mathbf{W}^{(1)} \rrbracket \right) + \vectorize \left( \tilde{\mathbf{X}} \right) \right) \\
        &= \matricize \left( \vectorize \left( \llbracket \tilde{\mathbf{X}}; \mathbf{S}^{(1)}, \mathbf{W}^{(1)} \rrbracket + \tilde{\mathbf{X}} \right) \right) \\
        &= \llbracket \tilde{\mathbf{X}}; \mathbf{S}^{(1)}, \mathbf{W}^{(1)} \rrbracket + \tilde{\mathbf{X}}
    \end{aligned}
\end{equation}
This corresponds to the forward pass in (\ref{eq:gtn_general}) with a pre-processed input, $\tilde{\mathbf{X}} = \textbf{W}^{(x)} \textbf{X}^T$, and a skip-connection. Figure \ref{fig:gtn_rnn} illustrates (\ref{eq:derivation_gtn_rnn}) in the TN notation.

\begin{remark}
The GSO, $\textbf{S}^{(1)} \in \mathbb{R}^{I_1 \times I_1}$, of an RNN can be interpreted as the graph adjacency matrix of a time-domain graph, where the $I_1$ vertices represent consecutive time-steps, while an edge connecting two time-steps has an edge weight that decays as a function of their time-gap, $s^{(1)}_{i_1, i_2} = c^{ i_1 - i_2 }$ for $i_1 > i_2$. Therefore, the further away two time-steps are, the weaker the connection between them. The condition $s_{i_1, i_2} = 0$ for $i_1 \leq i_2$ ensures the directed flow of time, where the past time-steps can influence futures states, but not vice-versa.
\end{remark}

\begin{remark}
If we constrain $\textbf{W}^{(1)} = \textbf{I}$, then the recurrent computation of hidden states in (\ref{eq:rnn_full}) simplifies to $\textbf{y}_{i_1} = c \textbf{y}_{i_1-1} + \textbf{W}^{(x)} \textbf{x}_{i_1}$, which simplifies (\ref{eq:derivation_gtn_rnn}) to the un-normalized GCN forward pass as
\begin{equation} \label{eq:derivation_gtn_rnn_simp}
    \begin{aligned}
        \textbf{Y}
        &= \llbracket \tilde{\mathbf{X}}; \mathbf{S}^{(1)}, \textbf{I} \rrbracket + \tilde{\mathbf{X}} \\
        &= \mathbf{S}^{(1)} \tilde{\mathbf{X}} + \tilde{\mathbf{X}} \\
        &= \left( \textbf{I} + \mathbf{S}^{(1)} \right) \tilde{\mathbf{X}}
    \end{aligned}
\end{equation}    
\end{remark}

\section{Beyond Classical Architectures} \label{sec:beyond}

This section discusses different ways in which the proposed framework enables the design of neural networks that improves upon the classical ones discussed in Section \ref{sec:classical_nnas}.

\subsection{The Tensorization Trick} \label{sec:tensorization_trick}

The classical neural network architectures discussed in Section \ref{sec:classical_nnas} are inherently based on ``flat-view" matrices and vectors. However, every component in the GTN framework can be readily tensorized to leverage the super-compression properties of Tensor Decomposition (TD) methods, which reduces the computational complexity of the underlying neural network architecture.

\begin{figure}[t]
    \centering
    \includegraphics[width=1.0\linewidth]{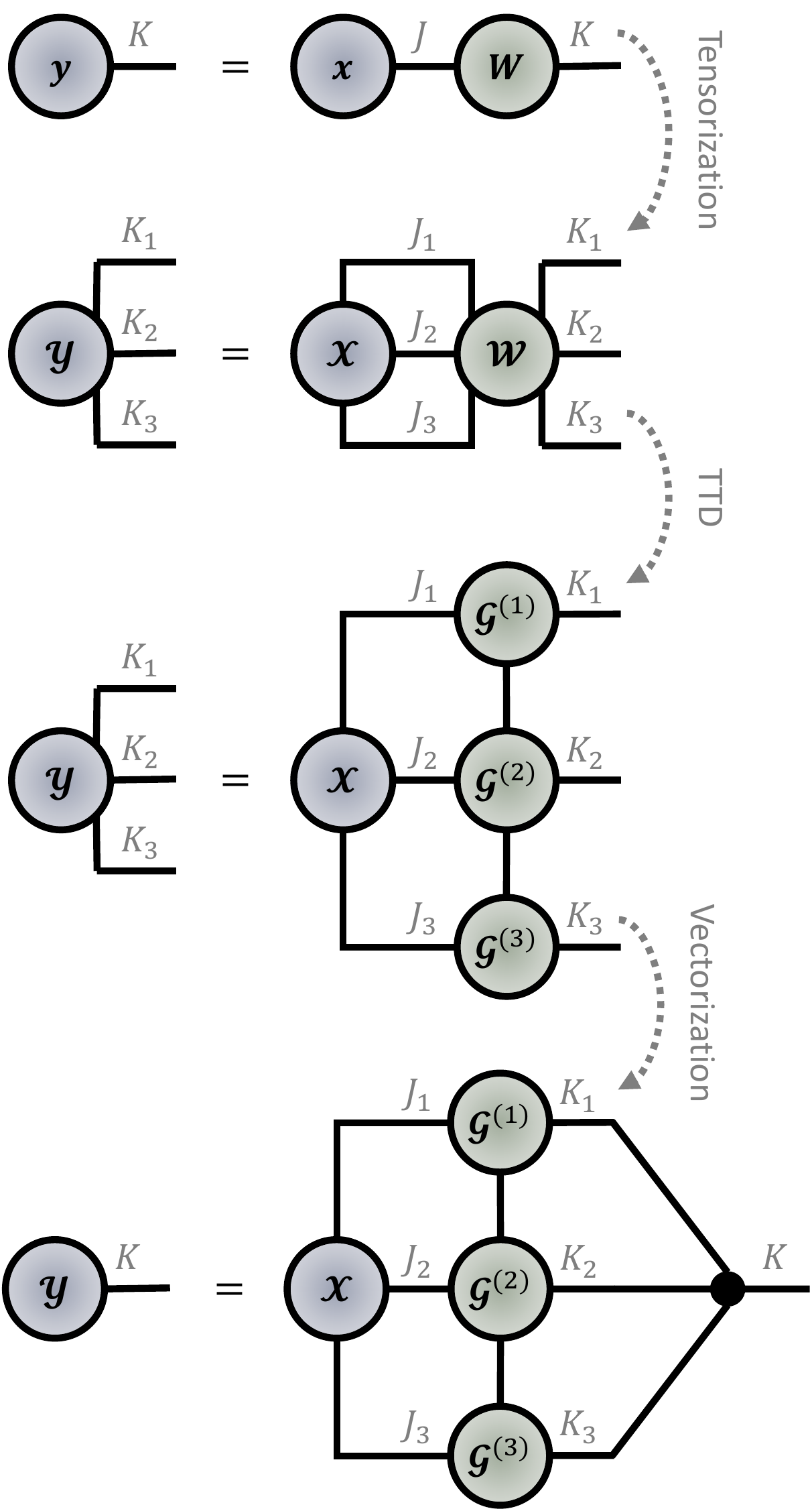}
    \caption{Weight matrix tensorization trick of order-$N=3$. The weight matrix is tensorized and stored in its low-rank Tensor-Train Decomposition (TTD) format.}
    \label{fig:wtd_trick}
\end{figure}

The authors in \cite{ttnn} were the first to tensorize and apply tensor decomposition to the weight matrix of DNNs. This was achieved by tensorizing the DNN forward pass from $\mathbf{y} = \mathbf{W} \mathbf{x}$ to $\mathcalbf{Y} = \mathcalbf{W} \times_{2,4,\ldots,2N}^{1,2,\ldots,N} \mathcalbf{X}$, where $\mathcalbf{Y} \in \mathbb{R}^{K_1 \times \cdots \times K_N}$, $\mathcalbf{W} \in \mathbb{R}^{(K_1 \times J_1) \times \cdots \times (K_N \times J_N)}$, and $\mathcalbf{X} \in \mathbb{R}^{J_1 \times \cdots \times J_N}$ are respectively the tensorizations of $\mathbf{y} \in \mathbb{R}^{K}$, $\mathbf{W} \in \mathbb{R}^{K \times J}$, and $\mathbf{x} \in \mathbb{R}^{J}$, such that $\prod_{n=1}^{N} J_n = J$ and $\prod_{n=1}^{N} K_n = K$. The weight tensor was then stored directly in its low-rank TTD format as $\mathcalbf{W} \approx \mathcalbf{G}^{(1)} \times^1_4 \cdots \times^1_4 \mathcalbf{G}^{(N)}$, where $\mathcalbf{G} \in \mathbb{R}^{R_{n-1} \times K_n \times J_n \times R_n}$ are TTD core tensors as defined in \ref{sec:ttd}. Figure \ref{fig:wtd_trick} illustrates the considered weight matrix tensorization trick in the TN notation, which allows for highly intuitive design of such neural learning systems.

\begin{example}
By storing the weights in a low-rank TTD form, the tensorization trick reduces the number of parameters from an original $JK = \prod_{n=1}^N J_n K_n$ (exponential in N) to $\sum_{n=1}^N R_{n-1} J_n K_n R_n$ (linear in N), which is highly efficient for small $R_n$ and large $N$. For instance, if the original feature vector is of size $J=256$ and the neural network has $K=256$ hidden units, then the weight matrix would have $65,536$ parameters. By performing the tensorization trick with $I_n=K_n=R_n=2$ and $N=8$, the number of parameters is reduced to $112$, resulting in a compression factor of $99.83\%$.
\end{example}

The considered tensorization trick is general and can be applied to any weight matrix of any neural network architecture using any tensor decomposition technique, such as Dense Neural Networks \cite{ttnn, calvi2019compression}, Convolutional Neural Networks \cite{sun2020deep}, Recurrent Neural Networks \cite{ttrnn, xu2021ttrnn}, and the Attention Mechanism \cite{xu2022attention}. Tensorized neural network models were shown to achieve comparable or better performance compared to their non-compressed counterparts while drastically reducing the associated complexity costs.


The tensorization trick can also be applied to GSOs of the GTN framework. For instance, the convolution tensor is known to have an exact QTTD representation \cite{Mandic2017}. Similarly, the computation of the attention based GSO can also benefit from the tensorization trick as shown in \cite{xu2022attention}. However, for highly irregular GSOs, an exact TD form might not exist and a large TD rank might be needed to approximate the original GSO matrix.


\subsection{Multi-Graph Modelling} \label{sec:multi_graph_modelling}

As discussed in Section \ref{sec:classical_nnas}, classical neural network architectures are designed to process data with only one domain mode. However, modern data sources often reside on multiple irregular domains, such as video data (pixel data over time), traffic data (time-series data over graph), and social network data (natural language data over graph). Designing a neural network architectures for processing data over multiple domains is particularly challenging, both in terms of handling multiple data dimensions jointly and digesting the sheer data volume. Using classical neural network architectures, this will likely require a combination of RNNs, CNNs, and GNNs that will inevitably explode in size due to the high-dimensionality of the problem. However, the considered data can be readily handled under the proposed GTN framework, which bypasses the cumbersome handling of multiple dimensions through intuitive TN diagrams. In addition, it can leverage the tensorization trick to effectively bypass the associated Curse-of-Dimensionality of the underlying problem. The design of GTN over multiple graph domains is next illustrated in Section \ref{sec:experiments}.

\section{Experiments} \label{sec:experiments}

To illustrate the flexibility of the proposed GTN framework for designing neural learning systems on multiple graph domains, we considered three different tasks of time-series modelling on graphs. 

\subsection{Datasets}

\begin{figure}
    \centering
    \includegraphics[width=1.0\linewidth]{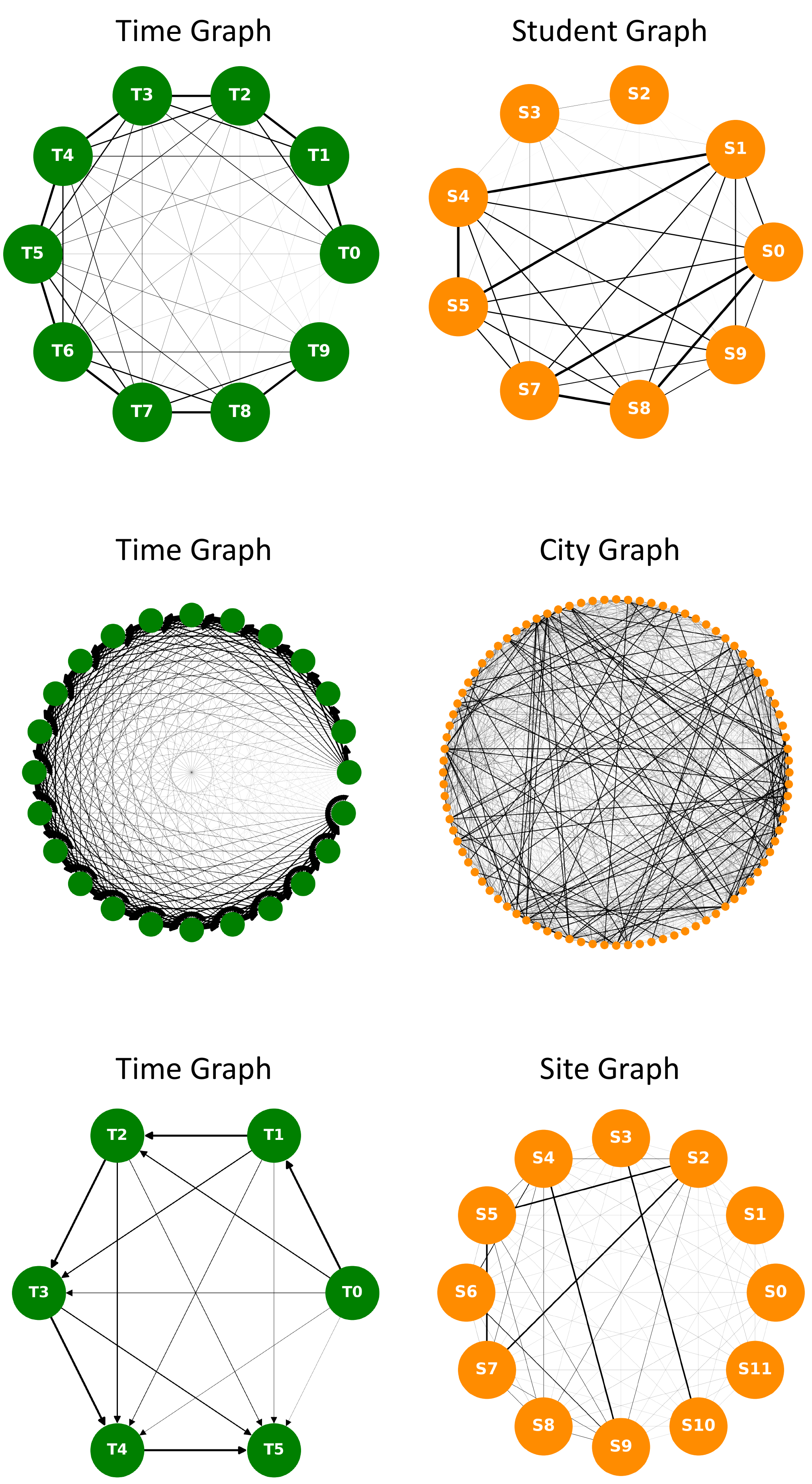}
    \caption{Graphs associated with the EEG classification (top), Temperature Forecasting (middle), and Air-Quality Forecasting (bottom) experiments. The time-domain graph is coloured in green, while the task-specific graph domain is colored in orange.}
    \label{fig:exp_graphs}
\end{figure}

\subsubsection*{EEG classification} Considers the classification of student mental states (confused or not) from their electroencephalogram (EEG) readings as they watch online education videos \cite{wang2013using}. The experiment data consists of $J_1=11$ EEG time-series features indexed over $I_1=10$ time-steps and recorded across $I_1=9$ students. The associated graphs domains are (i) a time-domain graph as discussed in Section \ref{sec:rnn_as_gtn}, and (ii) a student graph where the vertices represent students and the edges encode the demographic similarity between pairs of students.

\subsubsection*{Temperature Forecasting} Considers the predictive modelling of monthly temperature levels across cities in United States \cite{rohde2013new}. The experiment data consists of $J_1=3$ temperature variables indexed over $I_1=24$ time-steps and recorded across $I_2=92$ cities. The associated graphs domains are (i) a time-domain graph as discussed in Section \ref{sec:rnn_as_gtn}, and (ii) a city graph where the vertices represent cities and the edges encode the geographical proximity between pairs of cities.

\subsubsection*{Air-Quality Forecasting} Considers the predictive modelling of PM2.5 level across 12 different sites in Beijing \cite{zhang2017cautionary}. The experiment data consists of $J_1=27$ air-quality variables indexed over $I_1=6$ time-steps and recorded across $I_2=12$ different sites in Beijing. The associated graphs domains are (i) a time-domain graph as discussed in Section \ref{sec:rnn_as_gtn}, and (ii) a site graph where the vertices represent sites and the edges encode the geographical proximity between pairs of cities.

\subsection{Models}

\begin{figure}[h]
    \centering
    \includegraphics[width=1.0\linewidth]{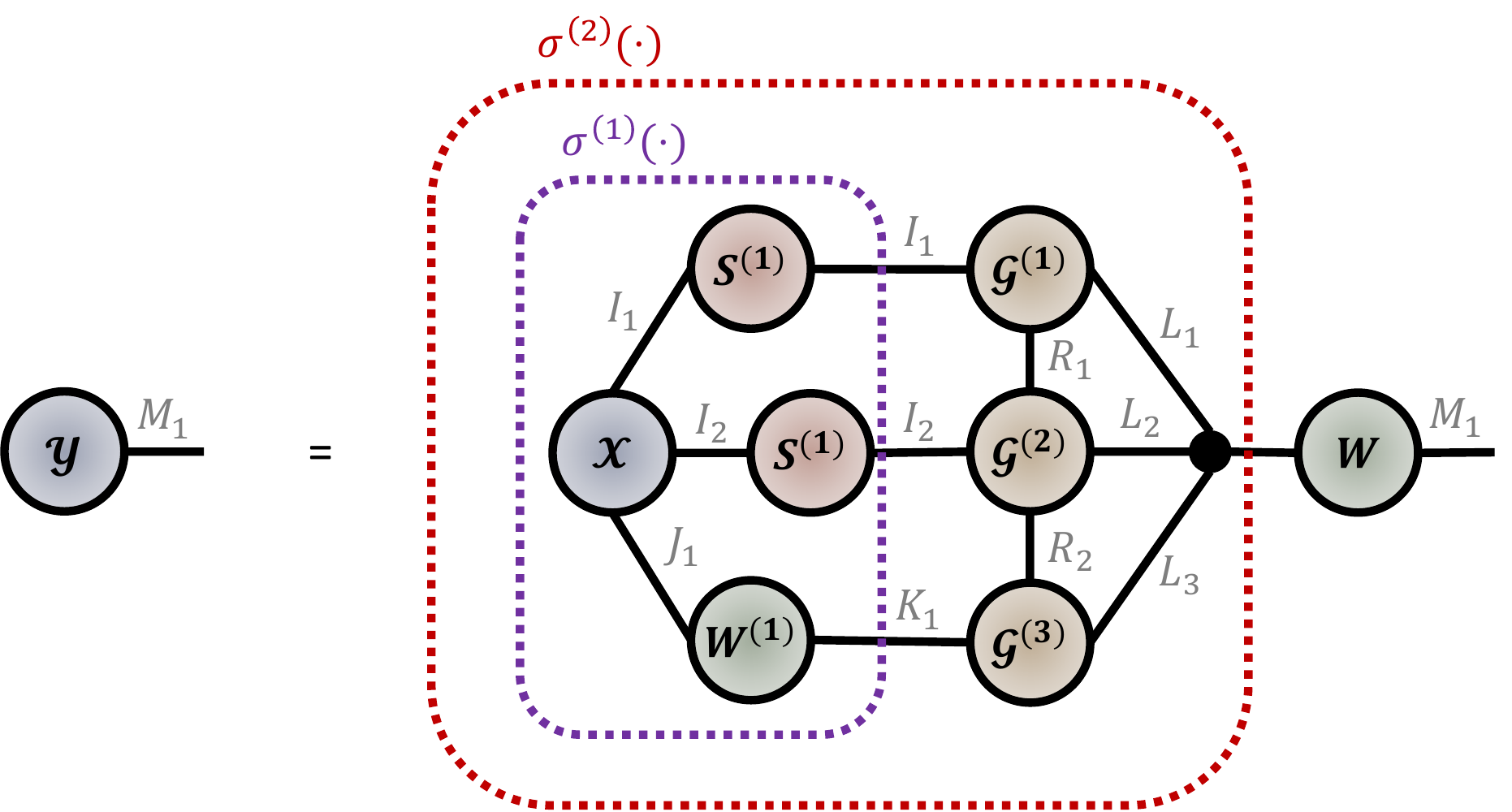}
    \caption{Neural network architecture designed to process the order-$(2+1)$ data tensor considered in the experiments. It contains: (i) A GTN layer with a time-domain GSO, $\mathbf{S}^{(1)}$, a task-specific GSO, $\mathbf{S}^{(2)}$, and a trainable weight matrix, $\mathbf{W}^{(1)}$; (ii) an activation function, $\sigma^{(1)}(\cdot)$; (iii) A DNN layer in a low-rank TTD form with trainable cores, $\mathcalbf{W} = \mathcalbf{G}^{(1)} \times_4^1 \mathcalbf{G}^{(2)} \times_4^1 \mathcalbf{G}^{(3)}$; (iv) an activation function, $\sigma^{(2)}(\cdot)$ (v) a vectorization operation; (vi) a standard DNN layer with a trainable weight matrix, $\mathbf{W}$.}
    \label{fig:exp_architecture}
\end{figure}

For all considered datasets, the input consisted of an order-$(2+1)$ data tensor (with $N=2$ domain modes corresponding to the time-mode and graph-mode, and $M=1$ feature mode), $\mathcalbf{X} \in \mathbb{R}^{I_1 \times I_2 \times J_1}$, containing $J_1$ descriptive features indexed over $I_1$ time-steps and across $I_2$ vertices in task-specific graphs. For the considered data tensor, a neural network architecture can be straightforwardly designed using the proposed GTN framework, as illustrated in Figure \ref{fig:exp_architecture}. The associated time-domain GSO, $\mathbf{S}^{(1)} \in \mathbb{R}^{I_1 \times I_1}$, is designed as discussed in Section \ref{sec:rnn_as_gtn}, while the task-specific GSO, $\mathbf{S}^{(2)} \in \mathbb{R}^{I_2 \times I_2}$, is designed using the corresponding adjacency matrix as discussed in Section \ref{sec:gtn_gnn}.

To illustrate the advantages of the GTN framework, we compare the proposed model against its \textit{equivalent} standard RNN and GCN architectures. More specifically, we define: 
\begin{itemize}
    \item A RNN architecture where: (i) the GTN layer was replaced by a classical RNN layer, and (ii) the DNN layer in TTD format was replaced by a standard DNN layer. The associated input data tensor was matricized as $\textbf{X}_{(1)} \in \mathbb{R}^{I_1 \times I_2 J_1}$ to accommodate the time-mode only structure of the RNN. 
    \item A GCN architecture where: (i) the GTN layer was replaced by a classical GCN layer, and (ii) the DNN layer in TTD format was replaced by a standard DNN layer.  The associated input data tensor was matricized as $\textbf{X}_{(2)} \in \mathbb{R}^{I_2 \times I_1 J_1}$ to accommodate the graph-mode only structure of the GCN. 
\end{itemize}

\subsection{Results}

The experiment results are summarized in Table \ref{tab:exp_res}, which includes performance metrics such as classification accuracy for the EEG classification experiment (EEG), and Mean-Squared-Error (MSE) for the Temperature (Temp) and Air-Quality (Air) forecasting experiments. The total number of trainable parameters (Params) is also reported as a measure of the model space complexity. The proposed GTN model outperformed all other considered models across all experiments and performance metrics, while using only a fraction of trainable parameters. This is achieved by virtue of its tensor structure, which can readily process large- and multi-dimensional data on multiple irregular domains jointly, and at low complexity costs. The classical RNN and GCN architectures suffered from the \textit{Curse-of-Dimensionality}, as they could only process one domain at a time, which drastically increased the required number of trainable parameters, leading to worse out-of-sample performance.

\begin{table}[h]
\begin{center}
    \begin{tabular}{l l l l l}

        \toprule
        &  & GTN & RNN & GCN \\

        \midrule
        EEG  & Accuracy & \textbf{53.67\%} & 50.98 \% & 46.84 \% \\
             & Params & \textbf{585} & 5055 & 3103 \\
        
        \midrule
        Temp & MSE & \textbf{0.0188} & 0.0438 & 0.0272 \\
             & Params & \textbf{1894} & 36740 & 84395 \\
    
        \midrule
        Air  & MSE & \textbf{0.1321} & 0.1680 & 0.3362 \\
             & Params & \textbf{486} & 8516 & 2180 \\

        \bottomrule

    \end{tabular}    
\end{center}
\caption{Experiment Results.}
\label{tab:exp_res}
\end{table}

\section{Conclusion} \label{sec:conclusion}

We have introduced a graphical method for describing and designing neural networks through a joint consideration of graphs and tensor networks. The so introduced Graph Tensor Network (GTN) framework has been shown to be general enough to include many modern neural network architectures as special cases, while being flexible enough to allow for the processing of data over any and many domains. The power and flexibility of the proposed framework has been demonstrated through several real-data case studies. It is our hope that by offering a common graphical language to formally describe neural networks, this will open avenues for including sophisticated tensor techniques developed in other fields, such as quantum many-body physics, for designing new classes of neural networks with superior performance and complexity characteristics.






%

\bibliographystyle{IEEEtran}
\bibliography{IEEEtran/references}




\end{document}